# The Silent Problem - Machine Learning Model Failure - How to Diagnose and Fix Ailing Machine Learning Models

Michele Bennett, PhD, Jaya Balusu, Karin Hayes, Ewa J. Kleczyk, PhD

April 22, 2022

## ABSTRACT

The COVID-19 pandemic has dramatically changed how healthcare is delivered to patients, how patients interact with healthcare providers, and how healthcare information is disseminated to both healthcare providers and patients. Analytical models that were trained and tested pre-pandemic may no longer be performing up to expectations, providing unreliable and irrelevant learning (ML) models given that ML depends on the basic principle that what happened in the past are likely to repeat in the future. ML faced to two important degradation principles, *concept drift*, when the underlying properties and characteristics of the variables change and *data drift*, when the data distributions, probabilities, co-variates, and other variable relationships change, both of which are prime culprits of model failure. Therefore, detecting and diagnosing drift in existing models is something that has become an imperative. And perhaps even more important is a shift in our mindset towards a conscious recognition that drift is inevitable, and model building must incorporate intentional resilience, the ability to offset and recover quickly from failure, and proactive robustness, avoiding failure by developing models that are less vulnerable to drift and disruption.

### Declaration

- Authors are affiliated with Symphony Health, a division of Icon, plc.
- Dr. Kleczyk is also an Affiliated Graduate Faculty in the School of Economics at the University of Maine, Orono, Maine
- Dr. Bennett is also an Adjunct Professor, within the Graduate and Doctoral Programs in Data Science, Computer Science, and Business Analytics at Grand Canyon University
- Competing Interest: The authors declare that they have no competing interests.
- Funding: Authors work for Symphony Health, ICON plc Organization.
- Author contact: Michele Bennett, mbennett1107@gmail.com

## Our World Has Changed

The COVID-19 pandemic has dramatically changed how healthcare is delivered to patients, how patients interact with healthcare providers, and how healthcare information is disseminated to both healthcare providers and patients (Mehrotra et al., 2020).  As a result of the structural changes brought on by the pandemic in the healthcare industry, many strong performing analytical models that were trained and tested pre-pandemic may no longer be performing up to expectations, and therefore, provide unreliable and irrelevant suggestions for implementation (Duckworth et al, 2021; Heaven, 2020 May 11).  Machine learning models depend on the basic principle that what happened in the past are likely to repeat in the future, and that there is a semantic memory much in the way that humans recall information from their past memories  (Gershman, 2017).  In a world potentially irrevocably changed, machine learning models faced to two important degradation principles,  *concept drift* and *data drift* (Babic et al., 2021 February; Duckworth et al., 2021).   Concept drift occurs when the underlying properties and characteristics of the target or dependent variables of the machine learning model change (Babic et al., 2021 February). Data drift occurs when the data distributions, probabilities, co-variates, and other variable relationships change (Duckworth et al., 2021).  Both concept and data drift contribute to model failure.

## Model Impact of Concept and Data Drift

Concept and data drift are not new and are risks that machine learning models do face, especially in dynamic environments (Duckworth et al., 2021) but in the advent of COVID-19, shifts in global behaviors and long-standing assumptions occurred almost overnight and assumed relationships between variables became instantly invalid (Isaksson, 2020 June 3).  For example, Figure 1 shows the count of NYC vehicle accidents from 2017-2020 from data from the NYPD.

Collisions followed a pattern that would allow for prediction analyses until March 2020, when suddenly fewer cars we on the road.

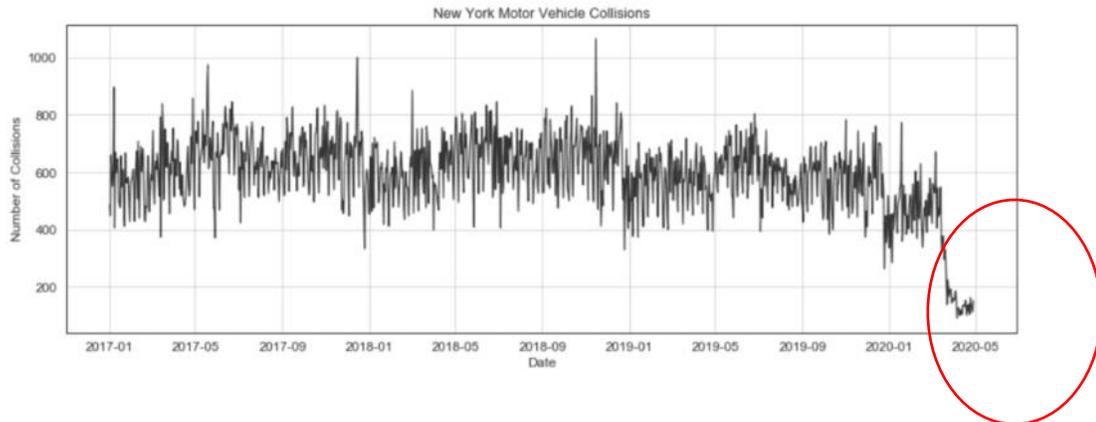

Figure 1: New York City motor vehicle collisions from 2017 to 2020 generated from publicly available NYPD data (Isakson, 2020 June 3).

In retail, Instacart's online grocery shopping services use machine learning models to predict product availability at stores, and prior to March 2020 enjoyed a 93% accuracy rate, which shrank to 61% after as customer dramatically changed shopping patterns as we all remember the runs on toilet paper and other staples (Deshpande, 2020 December 22). Image recognition software models at Getty Images started struggling with the concepts of "work", "play", "office", and "home" as people depicted surrounded by toys and kids redefined our norms as in Figure 2 (Gagliano, 2020 August 2).

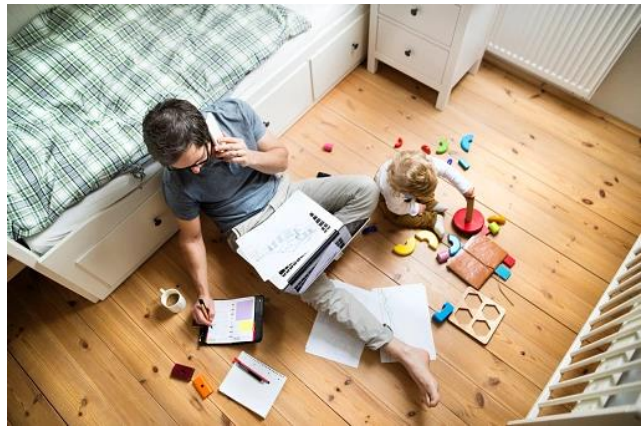

Figure 2: Man working from home. Westend61/Getty Images (Gagliano, 2020 August 2)

Even in predicting COVID-19 infections and impacts, many models were deployed but rarely were they successful (Heaven, 2021 July 30).  Wynants et al. (2020) reviewed hundreds of papers and predictive COVID-19 models that aimed to predict many pandemic scenarios including, mortality risk, progression to severe disease, intensive care unit admission, ventilation, intubation, or length of hospital stay.  Most models were deemed to have high data bias including non-representativeness of data, overfitting, and data drift from real-world clinical settings that were poorly understood (Wynants et al., 2020).   In studies of consumer sentiment concerning COVID-19 vaccines, Muller and Salathe (2020) found that models that were trained pre-COVID-19 vaccine approvals missed the decline in sentiment concerning vaccines due to concept drift.  Figure 3 below shows that models trained before vaccine approvals misclassified social media comments and performance declined 20% and quickly due to concept drift (Muller & Salathe, 2020).

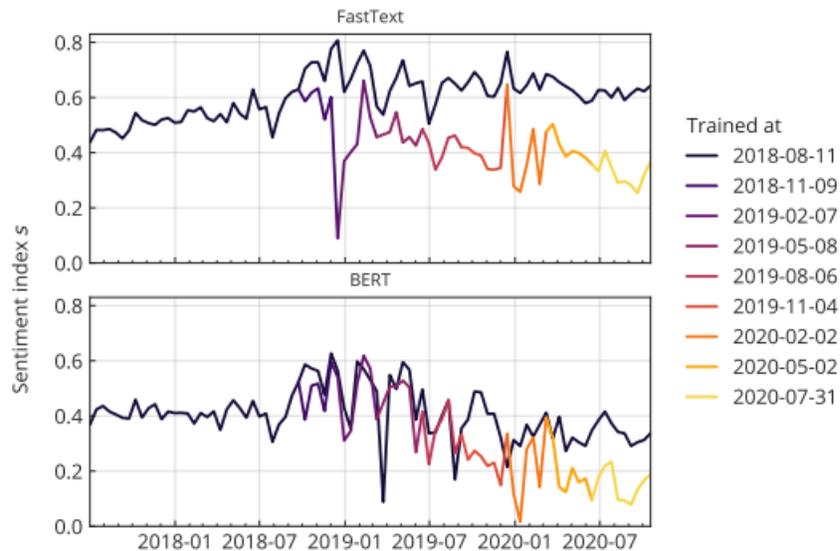

Figure 3: Social media sentiment performance surrounding COVID-19 vaccines demonstrating concept drift depending on when the models were trained (Muller & Salathe, 2020).

### The Future will Remain Unclear

With the future of the pandemic state still being unclear, many researchers and practitioners are advocating new discussion on resilience, robustness, sustainability, and risk (Galaitsi et al., 2021). It may not be enough to think about model building for resilience, as resilience is the ability of systems to recover from disruption, therefore a system that can withstand disruption without failure is considered robust (Urquiza et al., 2021). The focus on resilience and robustness represents newer thinking in the world of climate risk, and the predictive dimensions – flexibility/persistence, memory/learning, self-transformation, governance, and subject matter relevance (Urquiza et al., 2021) are transferrable to healthcare machine learning modeling to instill resilience and robustness. In order to get to this new normal, healthcare and pharmaceutical organizations should focus on short-term planning in order to develop models that are designed with risk, resilience, and robustness at the start, and on the identification of both concept and data drift in order to learn how to build upon this new way of model development until they reach a level of experience and confidence (Babic et al., 2021; Datta, 2021 Jan 27; Galaitsi, et al., 2021; Simion, 2020). Therefore, the mantra of using as much of historical data as possible to develop long-term analytical models cannot be supported in the current environment (Simion, 2020). Therefore, learning to identify risk and potential drift as part building new resilient and robust models seems an imperative over long-lasting models (Datta, 2021 Jan 27; Galaitsi et al., 2021, Lu et al., 2020), given the extreme evidence of the potentially fragility of many existing models (Heaven 2020 May 11).

### Identifying Models that are Prone to Drift and Failure

Healthcare modeling has a special imperative to focus on resilience and robustness given the potential impacts to the lives of patients so understanding the models and scenarios ripe for drift are some of the initial lessons needed (Chakravorti, 2022 March 17).

**Data Volatility and Data Drift**

Multiple and disparate datasets with little governance across sources, different levels of granularity, collection methods that are not specific for model use, and consistently evolving data features and dimensions, present reasons for data drift and therefore model failure (Babic et al., 2021 February; Chakravorti, 2022 March 17; Datta, 2021 Jan 27).  At the onset of COVID-19, Boston Children's Hospital and the Canadian news scraper, BlueDot, teamed together to predict cities at risk for travel weeks before WHO was able to even react, thereby having a good bit of success and perhaps setting the stage for the expectations of modeling to come (Chakravoti, 2022 March 17). However subsequent efforts to diagnose COVID-19 from lung scans and coughing sounds were not successful largely to what the authors called, *Frankenstein datasets*, or data compiled from disparate, multi-composite, and repackaged sources not all collected for the purposes being used and at different levels of granularity (Roberts et al., 2021). Sparse data that shifted directions almost as fast as it was collected, also contributed to otherwise stable models in predicting patient admission to ER, necessitating not just the need to update models but also choose different approaches moving from a simple classifier to ensemble modeling with explainable methods to not just identify when data drift happened but the impact to the model as soon as drift started happening so that ERs could be better prepared (Duckworth et al., 2021). For the explainable methodology, SHAP Additive Explanations, an additive feature attribution approach (Lundberg & Lee, 2017),  was used to monitor and identify drift, understand the most important individual variables in predictions, and monitor the changes in variable importance without invalidating the modeling efforts or even performance because the approach can measure the change in features importance relative to the target variable (Duckworth et al., 2021). Thereby seeing data drift and monitoring impact before failure or deep dip in performance.

## Environment and Assumption Volatility and Concept Drift

Concept drift is as prevalent as data drift as concept drift happens when the relationship between predictor and predicted variables change (Duckworth et al., 2021).  For example, a diagnostic algorithm that uses images of patients' skin to detect skin cancer failed in summer when more people with more sun exposure have a different color tone from the skin tones used in training data (Babic et al., 2021 February).  In addition, the incidence of patients relocating across the US during COVID-19 changed some of the geographic characteristics of incidence and seasonality for asthma, allergies, and other conditions (Heaven, 2020 May 11).

## Testing for Drift

If a model is being run consistently, such as the one in Figure 1, drift possibility would emerge as was seen in the significant drop in collisions almost overnight (Isakson, 2020 June 3). With the gift of hindsight, the fact that almost overnight drivers disappeared from the road was obviously the main reason behind the plunge. However, drift is likely to be less precipitous, potentially less obvious, and take more time to appear (Isakson, 2020 June 3).  Using a test such as the Dickey-Fuller test, a statistical test to determine if a time series is either non-stationary or stationary (or at least trend-stationary) (Jalil et al., 2019). If the time series is non-stationary, the change could be due to concept drift, especially if the change is not typically expected along seasonal trends (Jalil et al., 2019).   Therefore, testing should become part of regular model maintenance and if model maintenance is not part of standard operating procedures, COVID-19 has taught us that perhaps it should be.

## Identifying Drift Potential

Rather than waiting until drift is generating issues in performance, consistent monitoring, albeit expensive, is likely to be able to identify issues before failure.  For example, shifts to telemedicine, reduced hospitalizations, and increases in laboratory utilization during COVID-19 caused performance degradation in models relying on claims and EHR data to predict resource utilization of high-risk patients due to shifts in feature importance, dimension inputs, and patient characteristics (Parikh et al., 2022).   Therefore, awareness of and monitoring not just models but the surrounding assumptions, environment, case mix, and treatment patterns play a significant role in the impact of drift on model performance (Parikh et al., 2022).

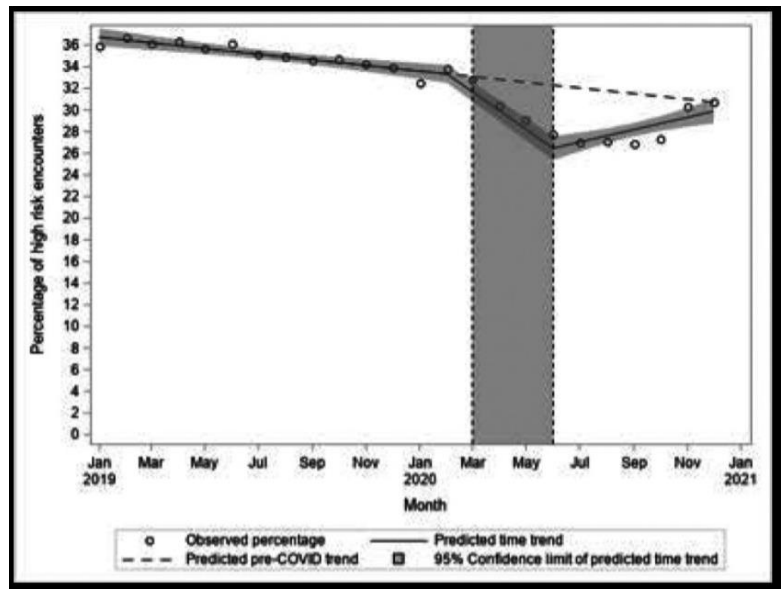

Figure 4: Performance drift in predict high-risk oncology patients and mortality risks before and
after COVID-19 showing a drop in true positive rates in the rates in grey ((Parikh et al., 2022).

## Models Prone to Drift

Therefore, there are circumstances, scenarios and external factors in healthcare that can be the source of drift, such as at the launch of a new product in a therapeutic area especially one in a category or novel approach where there are no trends or adoption patterns (Lu et al., 2020).

Patient case mix changes (such as demographic or geographic changes), new or shifts in provider methods (such as increases in telehealth utilization, increase use of urgent care, or decreases in elective surgeries), access changes (such as coverage changes), and when the diversity of the market changes as companies try to address healthcare disparities (Babic et al, 2021 February; Duckworth et al., 2021).

There are model types such as those that are heavily time dependent or seasonal, are sensitive to missing values, susceptible to label leakage (i.e., when target is inadvertently or indirectly in the training data causing the model of over predict and minimize errors), and/or are based on highly volatile data (Babic et al., 2021 February; Duckworth et al., 2021; Lue et al., 2020; Parikh et al., 2022). Of course, in the evolving world of healthcare analysis, these could explain drift risk for any model. And while is it prudent to expect and test for drift in modeling, there is a growing movement in building models that are more resilient and can return to normal performance post a disruption, or more robust and can maintain performance during a disruption (Galaitsi et al., 2021).

## Building a Future for Resilience and Robustness

Detecting and diagnosing drift in existing models is something that has become of critical importance based on the experiences of disruption caused by COVID-19 (Babic et al., 2021 February; Duckworth et al., 2021). However, knowing the current performance and expectations can be a launch points as can be knowing the magnitude of current challenges.

### Observing Current Performance

While "do nothing" may be perceived as a weak choice, it can be a baseline for recognizing problems in existing models and provide a critical look at current performance against expectations as valid starting point. And part of that starting point, is asking the right questions of the model purpose, practical application, clinical importance, and fit for purpose (Bachtiger et al., 2021 February). As COVID-19 threw the healthcare system into crisis, it offered a great deal of opportunity to create new models using the enormous amount of data being generated to predict disease spread and impact but with little success (Heaven, 2021 July 30). Given the speed of the development of ML models, it is tempting to throw all the variables into the pool and push the button, which can yield interesting but not necessarily clinically usable results (Bachtiger et al., 2020).  Clinicians are trained on a fuller patient picture than may be able to be represented in models so a clinical perspective and critical review of the modeling assumptions, feature importance, and even some data that otherwise would be considered noise without a medical viewpoint are important considerations in model build (Bachtiger et al., 2020). A multi-functional team, including those with disease-specific subject matter expertise, can review label and feature engineering strategies and prioritize potential actionable insights and practical application, not just feature importance (Bachtiger et al., 2020).  This approach may yield more enthusiasm for the work of maintenance as clinicians and patients reap more benefit from the model outputs and outcomes.

**Diagnosing Strategies**

*Periodic Update*.  Depending on the importance of the model or complexity of the production system in which the model performs, it may be prudent to update model with more recent data, assume external conditions have shifted, or know that specific condition have changed such a new drug has launched that could change market share, a new diagnostic test is now available, or a new treatment for a rare disease is changing patient lives (Chakravoti, 2022 March 17; Isakson, 2020 June 3; Parikh et al., 2022).

Steps that can be taken to quantify drift in models include:

- Running Dickey-Fuller tests on time-series models to determine if the data are stationary and transformation of data to more stationary runs (Jalil et al., 2019)

- Measuring drift magnitude using Hellinger Distance to measure the distance between concepts (via distributions) at the start of modeling and the end of the period being measured, such as before and after COVID-19 (Hoens et al., 2012).

- Measuring drift duration which measures the elapsed time of the drift, which is important when studying drift patterns (Hoens et al., 2012). but less telling when the goal is to understand the presence of drift and immediately counterbalance.

*Feature Adjustment.*  Some ML models allow for weighting of data on input and that might be warranted when for example, false negatives in predicting cancer are more problematic than false positives or in device preventative maintenance when failures are potentially life-threatening so thereby more expensive than premature replacements.  However, feature weighting at input should be done with caution as it can create bias and allowing the data to weight features is one of the basic premises behind machine learning, and practitioners can prefer thresholds (Hart, 2019 Nov).

*Ensemble Balancing and Adaptive Modeling.*  Models that use continuous data streams tend to drift faster than those with more static data, but even static data can have "drifting" expectations and external factors whether they are gradual or abrupt (Sarnovsky & Kolarik, 2021).  Ensembles, which are popular due to the assumption that a set of classifiers together can achieve better performance than individual classifiers but given that the expectation that drift is more commonplace than ever, adaptive approaches to ensembles might not only detect drive but help deflect the impact (Sarnovsky & Kolarik, 2021).  Adaptive random forest and adaptive eXtreme Gradient Boosting methods are some of the methods that combine trees with drift detectors that sample with reposition rather than growing the trees sequentially and seek out warnings of drift and adjust (Gomes et al., 2017).

**Considering Resiliency in Model Development**

Resilience can be defined from different perspectives including one based on risk and vulnerability, whereas if there is a disruption such as was seen in COVID-19, resilience frames recovery in a perspective of tradeoffs between mitigating risk with enhancing resilience (Galaitsi et al., 2021). Vulnerability also can be a cumulative measure of the sensitivity of model components to drift but compounds the complexity of risk assessments  (Urquiza et al., 2021). Assessment of the risks or vulnerabilities to drift of the data utilized and the scenarios served by the model should be part of regular model checks with the frequency commiserate with the impact of model failure as part of an overall focus on resilience.

If resilience is set against efficiency, then strategies surrounding redundancy such as keeping stockpiles of PPE for example, or in terms of modeling, backup models or adaptive ensembles, with perhaps less efficient use of resources might be appealing to some organization and modeling purposes (Gomes et al., 2017).  Systems that run lean can be vulnerable to disruption and can fail catastrophically, while those that are more reliably resilient have built-in redundancies. For example, in general, patients at understaffed hospitals were less likely to survive during COVID-19 surges (Rosenthal et al., 2020).  Redundancy can limit vulnerabilities but increases costs and like using increased model checks to mitigate risk, the increase in costs and loss of efficiency should be weighed against the probability and impact of failure.

Recovery time is also part of the scope of consideration for resilience, and short-term vs. long-term recovery, which can be factored into the choices and scale of resiliency and again weighted by the impact of failures as well as time to recovery (Urquiza et al., 2021).

## Creating Robust Models and Robust Thinking

Perhaps a stronger approach is to consider proactive and robust model development. Robustness differs from resilience in that robust models maintain performance rather than recover when faced with potential disruptions or drift (Galaitsi et al., 2021; Urquiza et al., 2021). While resiliency is an approach that cannot be dismissed and should be part of all ML modeling frameworks since catastrophic shifts are harder to plan for and more disruptive than one could anticipate before the last two years (Galaitsi et al., 2021), they are nonetheless reactive and not proactive methods. Therefore, a proactive approach that is less vulnerable to disruption and drift should become the primary mindset (Luo et al, 2022), given the unprecedented challenges faced by machine learning across the world (Babic et al., 2021 February; Duckworth et al., 2021; Lue et al., 2020; Parikh et al., 2022).   But perhaps one of the most important lessons for modeling is the idea of intentionally building models with resilience and robustness from the start (Datta, A. 2021 Jan 27; Galaitsi et a., 2021; Urquiza et al., 2021).

## Specific Approaches for Healthcare Models

Robust and resilient healthcare models require specific approaches that provide the level of dynamic flexibility needed with intentional attention to stability of data, proactively sought clinical relevance,  use of real-world evidence, and creation of adaptive ensembles as shown in Figure 5 (Symphony Health, 2021).  Those approaches are:

- **Creation of data stabilization criteria** across fixed time periods, beyond eligibility criteria for cohort development, that is fit for purpose and less vulnerable to longitudinal drift (Hoens et al., 2012; Jalil et al., 2019; Symphony Health, 2021)

- **Performing feature engineering with clinical importance** that can incorporate thresholds to limit model bias (Hart, 2019 Nov) but also maintains clinical importance as a top priority (Bachtiger et al., 2020; Symphony Health, 2021).

- **Use of unsupervised & semi-supervised methods,** such as Hidden Markov Models, to find prognostic predictors as part of initial feature identification which then become part of semi-supervised learnings for full feature selection. Several studies have shown that use of semi-supervised methods with inclusion of HMM have boosted resilience in healthcare modeling (Symphony Health, 2021; Tamposis et al., 2019; Yoshida, 2021 August 30)

- **Intentional clinical verification and real-world testing** as part of development beyond feature engineering but also in preliminary outputs so that current clinical behaviors of providers and patients are tested with forward-looking real-world evidence (Bachtiger et al., 2020); Symphony Health, 2021.  This is a critical step in ensuring that the past really does reflect the future given the basic premise of predictive modeling (Gershman, 2017) that was one of the primary downfalls of pre-COVID-19 models (Babic et al. 2021 February; Duckworth et al., 2021). Forward-looking, real-world testing sets the data stability criteria to a future timeframe to test the model against actual outcomes to ensure that performance is robust, and drift is not a factor (Symphony Health, 2021).

- **Building adaptive ensembles** and testing methods that are cognizant that drift is likely and can identify and flex, combined with human oversight and insights from real-world testing (Sarnovsky & Kolarik, 2021; Symphony Health 2021).

The intentional approach described above addresses both resiliency and robustness with less loss of efficiency than approaches that focus on resilience alone by not be solely focuses on risk and redundancy (Gomes et al., 2017; Urquiza et al., 2021). Combining a focus on robustness through use of semi-supervised learning, real-world testing and periodic drift reviews, addresses both seeking out drift-damaging scenarios as well as limiting the impact should drift occur (Datta, A. 2021 Jan 27; Galaitsi et a., 2021; Symphony Health, 2021; Urquiza et al., 2021).

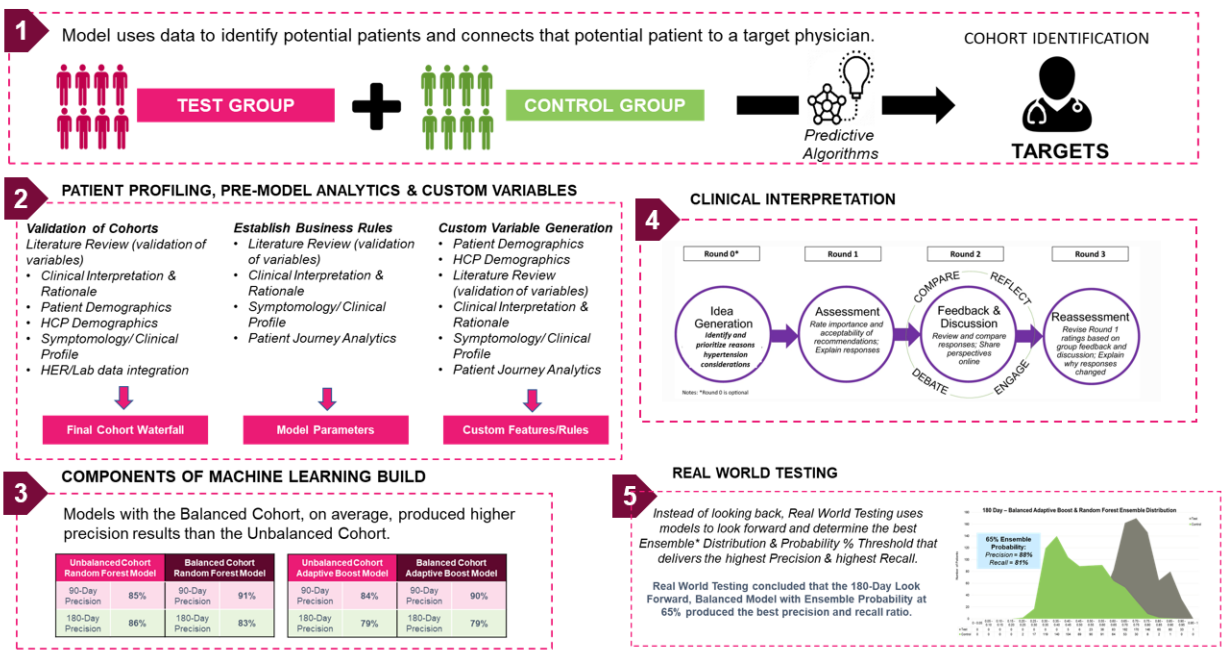

Figure 5: Robust and proactive healthcare machine learning model creation approach that incorporates clinical and subject matter expertise as part of modeling building and prospective testing to test future model performance in drift conditions (Symphony Health, 2021).

## Conclusion

The pandemic effect on machine learning models and AI was a severe example of the impact of both data and concept drift (Babic et al., 2021 February; Datta, 2021 Jan 27; Duckworth et al., 2021). Concept drift was due to the almost immediate change in the relationships between target and predictor variables and in the environment in which they interact (Babic et al., 2021 February). Data drift occurred due to extreme changes in probabilities, distributions, and operating assumptions.  (Duckworth et al., 2021).  Both concept and data drift contributed to widespread model failure in healthcare (Datta, 2021 Jan 27;

Isaksson, 2020 June 3).  Even models aimed at trying to predict COVID-19 outbreaks and impacts failed (Heaven, 2020 May 11; Wynants et al., 2020).   Therefore, many researchers, practitioners, and clinicians are looking toward a focus on healthcare model resilience and robustness as an intentional approach to decrease the model risk and increase recovery from drift,

and to even limit the overall vulnerability to drift, even of the extreme drift caused by COVID-19

(Babic et al. 2021 February; Duckworth et al., 2021; Symphony Health, 2021; Tamposis et al.,

2019; Yoshida, 2021 August 30).  By employing key resilience and robust model approaches to specifically, intentionally, and proactively address potential drift as a given consequence of using machine learning in the real world, model failure vulnerabilities can be identified, risks can be mitigated, and failure impacts reduced (Symphony Health, 2021).

by-predicting-the-progression-of-covid19-and-survival-of-patients-directly-from-their-

chest-ct-mages